%% file: main.tex
\documentclass[10pt,twocolumn,letterpaper]{article}

\usepackage[pagenumbers]{cvpr} 
\usepackage[accsupp]{axessibility}  

\definecolor{cvprblue}{rgb}{0.21,0.49,0.74}
\usepackage[pagebackref,breaklinks,colorlinks,allcolors=cvprblue]{hyperref}
\usepackage{multirow}
\usepackage{adjustbox}
\usepackage{verbatim}

\def\paperID{8381} 
\def\confName{CVPR}
\def\confYear{2025}

\title{Self-supervised ControlNet with Spatio-Temporal Mamba for Real-world Video Super-resolution}

\author{
Shijun Shi$^{1}$$^{*}$ \quad 
Jing Xu$^{2}$\thanks{Equal contribution.} \quad 
Lijing Lu$^{3}$$^{\dagger}$ \quad 
Zhihang Li$^{4}$\thanks{Corresponding author.} \quad 
Kai Hu$^{1}$\\
$^{1}$Jiangnan University \quad
$^{2}$University of Science and Technology of China \\
$^{3}$Peking University \quad
$^{4}$Chinese Academy of Sciences\\
{\tt\small ssj180123@gmail.com, xujing0@mail.ustc.edu.cn, lulijing1997@gmail.com,}\\
{\tt\small lizhihang.cas@gmail.com, hukai\_wlw@jiangnan.edu.cn}\\
{\small \url{https://ssj9596.github.io/scst-project/}}
}

\begin{document}
\maketitle
\input{sec/0_abstract}
\vspace{-0.5cm}
\input{sec/1_intro}

\input{sec/2_related_work}
\input{sec/3_Methodology}
\input{sec/4_experiments}
\input{sec/5_conclusion}
{
    \small
    \bibliographystyle{ieeenat_fullname}
    \bibliography{main}
}
\end{document}

%% file: sec/0_abstract.tex
\begin{abstract}
Existing diffusion-based video super-resolution (VSR) methods are susceptible to introducing complex degradations and noticeable artifacts into high-resolution videos due to their inherent randomness. In this paper, we propose a noise-robust real-world VSR framework by incorporating self-supervised learning and Mamba into pre-trained latent diffusion models. To ensure content consistency across adjacent frames, we enhance the diffusion model with a global spatio-temporal attention mechanism using the Video State-Space block with a 3D Selective Scan module, which reinforces coherence at an affordable computational cost. To further reduce artifacts in generated details, we introduce a self-supervised ControlNet that leverages HR features as guidance and employs contrastive learning to extract degradation-insensitive features from LR videos. Finally, a three-stage training strategy based on a mixture of HR-LR videos is proposed to stabilize VSR training. The proposed Self-supervised ControlNet with Spatio-Temporal Continuous Mamba based VSR algorithm achieves superior perceptual quality than state-of-the-arts on real-world VSR benchmark datasets, validating the effectiveness of the proposed model design and training strategies.
\end{abstract}

%% file: sec/1_intro.tex
\section{Introduction}
\label{sec:intro}
Video super-resolution (VSR) aims to restore high-resolution (HR) videos by leveraging complementary temporal information within low-resolution (LR) frames, which holds great value in practical usages, e.g., surveillance and high-definition display. Previous works mainly rely on the assumptions of simple and known image degradations (e.g., bicubic downsampling) or specific camera-related degradations, making it challenging to generalize the trained VSR models to real-world LR videos with unknown and more complex degradations.
\begin{figure}[t]
\centering
\includegraphics[width=1\linewidth]{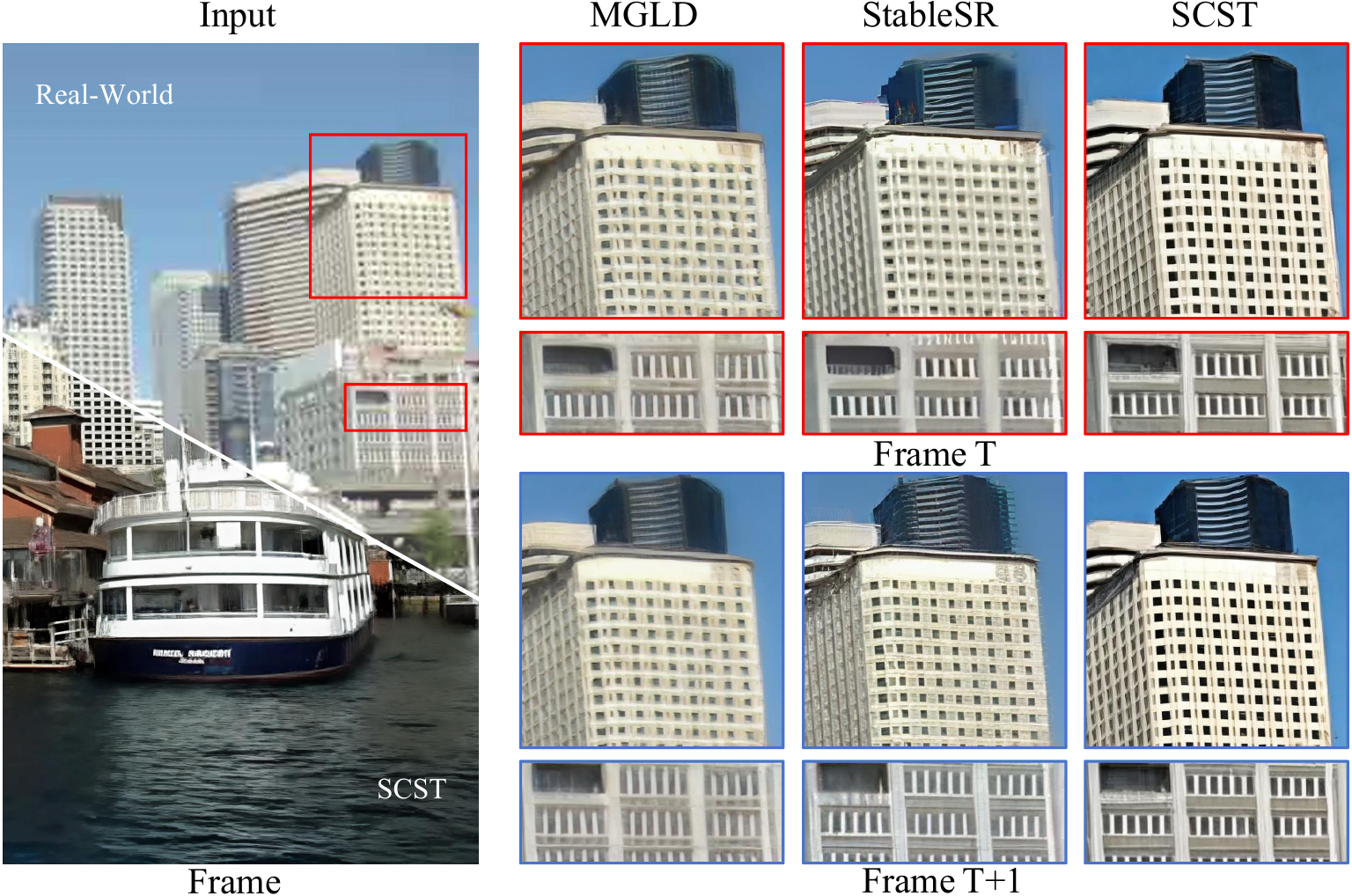}
\vspace{-0.55cm}
\caption{A side-by-side comparison of video super-resolution techniques StableSR~\cite{stablesr}, MGLD~\cite{mgld}, and our SCST on two adjacent frames from the VideoLQ dataset. The zoomed-in regions, captured from the same local position, illustrate each method's performance. SCST stands out for maintaining temporal consistency while delivering crisp details in real-world scenarios.}
\label{figure:motivation}
\vspace{-0.47cm}
\end{figure}

Recently, diffusion-based generative models \cite{ho2020denoising} have achieved great success in tasks of general image/video generation \cite{ho2022imagen,ramesh2022hierarchical,rombach2022high,saharia2022photorealistic} and downstream tasks such as image editing \cite{kawar2023imagic}, inpainting \cite{lugmayr2022repaint}, colorization \cite{carrillo2023diffusart}, etc, owing to its powerful ability in capturing diverse and complicated data distributions. 
There are also efforts to adapt these diffusion priors to image super-resolution \cite{kawar2022denoising,wang2024exploiting,yang2023pixel}. 
\cite{wang2024exploiting,yang2023pixel} rely on elaborately designed network architectures, such as ControlNet, to condition the diffusion models on low-resolution (LR) images for real-world image super-resolution. \citet{stablesr} propose to inject LR images into the U-Net blocks in a LDM network using a SFT module.
While showing promising results, directly applying pre-trained diffusion models for image super-resolution to
degraded videos is challenging due to the inherent randomness in diffusion sampling, leading to temporal inconsistencies in generated videos. To address this problem, \citet{upscale} propose to employ 3D convolution and temporal attention in the network to ensure temporal consistency.
\citet{mgld} introduce to use optical flow to align latent features between adjacent frames, enhancing temporal coherence.
However, the coupled problem of recovering the unknown and complex degradations in real-world scenarios and producing temporal consistent results in the same time causes great learning complexity. As shown in Figure \ref{figure:motivation}, while MGLD~\cite{mgld} focuses on temporal consistency, it exhibits shortcomings in spatial recovery.
Therefore, it is crucial for real-world VSR to extract clean features from complex degradation and hallucinate visual content while also modeling the spatial-temporal dependencies between different frames.

To tackle these issues, we propose a \textbf{S}elf-supervised \textbf{C}ontrolNet with \textbf{S}patio-\textbf{T}emporal Continuous Mamba (SCST) for real-world VSR. SCST is a noise-robust temporal coherence diffusion model, aimed at reconstructing fine-grained textures from videos with unknown degradations.
To our best knowledge, we are the first to introduce \textbf{S}patial-\textbf{T}emporal \textbf{C}ontinuous \textbf{M}amba (STCM) for global 3D attention in the VSR task. 
Specifically, we propose a 3D Selective Scan method tailored for traversing spatial-temporal domain, ensuring that every video patch acquires contextual knowledge through a compressed hidden state calculated along the respective scanning path with a linear computational complexity. 
To mitigate the impact of complex degradations in LR, we propose a ControlNet based on MoCo (MoCoCtrl) \cite{self-supervised_task6}, which can distill degradation-insensitive features from LR towards the HR target.
Finally, we design a multi-stage HR-LR hybrid training strategy to stabilize training. The self-supervised ControlNet is trained using contrastive learning, while the temporal module is optimized with de-noising loss. 

The main contributions of this paper are summarized as follows.
\begin{itemize}
\item We propose a noise-robust temporal coherence diffusion model based on self-supervised learning and spatial-temporal continuous Mamba. The proposed self-supervised ControlNet distills degradation-insensitive features from LR videos while a global 3D attention based on Mamba is designed to model the spatial-temporal relationship of the video.
\item To stabilize VSR training, we introduce a decoupled three-stage training strategy where HR and LR videos are mixed for training. Contrastive learning loss is incorporated to align LR features with HR, enabling the extraction of noise-free features.
\item Our proposed SCST model achieves state-of-the-art performance on existing benchmarks, showing remarkable visual realism and temporal consistency.
\end{itemize}

%% file: sec/2_related_work.tex
\section{Related Work}
\label{sec:related_work}
\subsection{Video Super-Resolution}
The goal of VSR is to enhance a sequence of HR video frames from their degraded LR counterparts. 
Based on the paradigms, existing VSR algorithms \cite{cao2021video,chan2021basicvsr,chan2022basicvsr++,isobe2020video,isobe2020video2,isobe2020revisiting,jo2018deep,liang2024vrt,liang2022recurrent,wang2019edvr,xue2019video,liu2013bayesian,nah2019ntire,yi2019progressive} could be roughly classified into two categories: temporal sliding-window based VSR and recurrent framework based VSR.
Temporal sliding-window based VSR \cite{jo2018deep,wang2019edvr,li2020mucan} utilize a fixed set of neighboring frames to super-resolve one or more target frames.
However,
the information accessible is constrained by the temporal
window’s size. Consequently, these methods can only exploit
the temporal details of a restricted subset of input video
frames.
To exploit temporal information from more frames, recurrent framework based VSR \cite{chan2021basicvsr,chan2022basicvsr++,liang2024vrt,liang2022recurrent} utilizes multiple LR frames as input and employs recurrent neural networks to simultaneously produce their corresponding SR results.
However, most existing approaches \cite{cao2021video,chan2021basicvsr,chan2022basicvsr++,isobe2020video,isobe2020video2,isobe2020revisiting,jo2018deep,liang2024vrt,liang2022recurrent,wang2019edvr,xue2019video} assume a pre-defined degradation process \cite{liu2013bayesian,nah2019ntire,yi2019progressive}. In real-world scenes with more complicated
degradations, these VSR methods may not perform well.
Due to the lack of real-world paired data for training, \citet{realvsr} propose to collect LR-HR data pairs with iPhone cameras to better model real-world degradations. 
While the VSR model trained on such data can be effective to videos captured by similar mobile cameras, it is relatively labor-intensive and may not generalize well to videos collected by other devices. 
Recent studies have shifted towards employing diverse degradations for data augmentation during training, such as blur, downsampling, noise and video compression \cite{realbasicvsr,xie2023mitigating}. 
However, maintaining temporal consistency while generating photorealistic textures remains a challenge. 

\begin{figure*}
    \centering
    \includegraphics[width=1\linewidth]{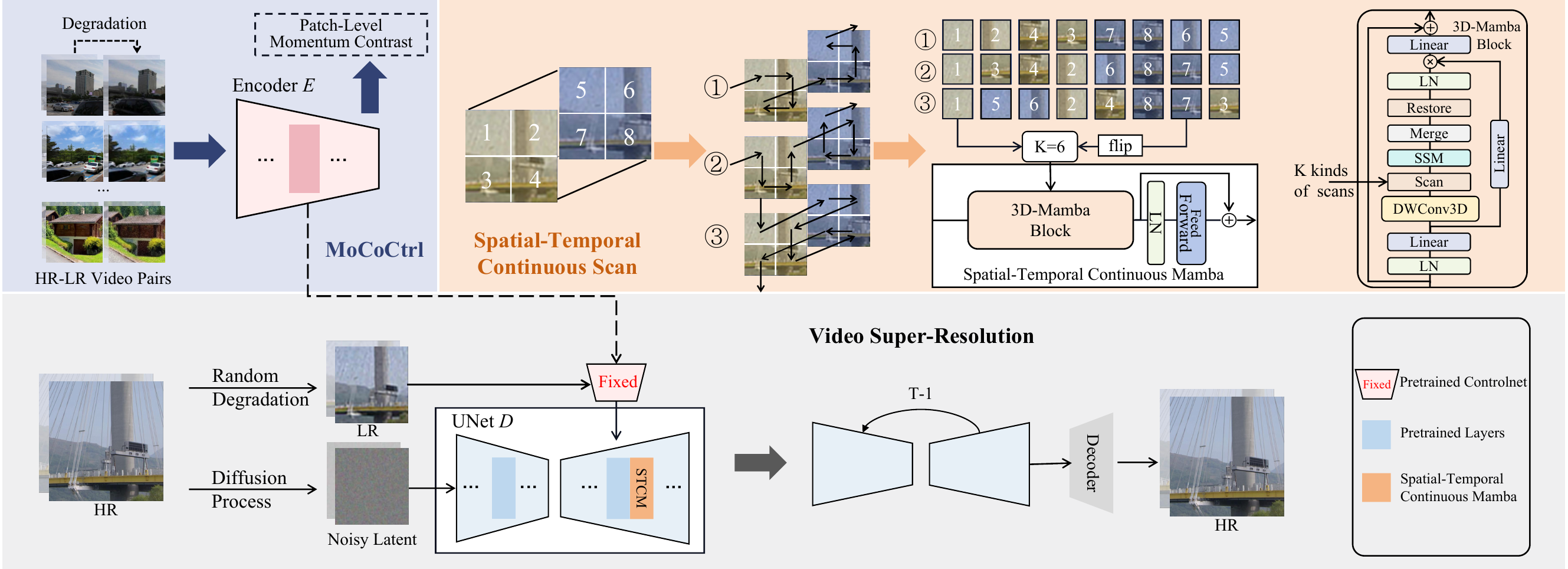}
    \vspace{-0.6cm}
    \caption{Overview of the proposed SCST framework for real-world VSR. SCST consists of several modules, including Spatial-Temporal Continuous Mamba (STCM) and Self-supervised ControlNet (MoCoCtrl). The STCM incorporates 3D-Mamba Block within its structure, which, with the addition of spatial-temporal continuous scan, ensures comprehensive 3D attention for both inter-frame and intra-frame modeling. The Self-supervised ControlNet adopts the MoCo architecture to employ contrastive learning between LR and HR features, aligning LR features to noise-free HR features, thus reducing the impact of degradation.}
    \label{fig:framework}
    \vspace{-0.5cm}
\end{figure*}

\subsection{State Space Models}
Structured state space models (S4) \cite{S4}, as a promising framework in handling long-distance sequences, has attracted widespread research interest. A variety of S4-inspired models, that capture long-range dependencies in sequential data, achieve competitive performance on various tasks \cite{variant1,variant2,variant3,variant4,variant5}. The major reason behind this might be that S4's adherence to Linear Time Invariance (LTI), which guarantees consistent output for identical inputs regardless of their temporal application. Nevertheless, LTI systems come with some limitations, especially when it comes to handling dynamic changes over time. The constancy of the internal state transition matrix throughout the sequence constrains the model's adaptability to evolving content, thereby limiting its utility in contexts demanding content-driven reasoning. 
To address these constraints, Mamba \cite{mamba} is recently introduced as a state-space model that dynamically adjusts its parameters in response to the input sequence. 
This adaptive strategy enables Mamba to engage in context-dependent reasoning, significantly enhancing its effectiveness across various domains \cite{domain1,domain2,domain3}. However, the application of Mamba in video super resolution tasks remains unexplored.

\subsection{Self-Supervised learning}
Self-supervised learning (SSL), as an off-the-shelf representation techniques, has achieved excellent performance in various computer vision tasks \cite{self-supervised_task1,self-supervised_task2,self-supervised_task3,self-supervised_task4,self-supervised_task5,self-supervised_task6,self-supervised_task7,self-supervised_task8}. Recently, 
contrastive learning \cite{cl1,cl2,cl3,self-supervised_task1,self-supervised_task2} has emerged as one of the most prominent self-supervised methods, making significant progress in exploring image representations based on instance discrimination tasks, where an instance’s different views originating from the same instance are treated as positive examples for an anchor sample, while views from different instances serve as negative examples. The core idea is to promote proximity between positive examples and maximize the separation between negative examples within the latent space, thereby encouraging the model to capture meaningful relationships within the data.

%% file: sec/3_Methodology.tex
\section{Methodology}
\subsection{Overall Architecture}
Given a LR video sequence of $T$ frames $x^l\in R^{T\times H\times W \times 3}$, the goal is to reconstruct the corresponding HR video sequence $x^h\in R^{T\times sH\times sW \times 3}$, where $s$ is the scaling factor and $H$, $W$ are the height and width of input frames. We build our method on top of a pretrained SD model \cite{rombach2022high} to harness the powerful generative priors for real-world VSR. The key design principle of the SD model is to add progressively increasing Gaussian noise to the clean data sample $x$ according to a noise schedule $\{\beta_t\}_{t=1}^T$. Given a noisy sample $x_t$ where $t$ is the diffusion step, a denoising UNet $D$ is trained to estimate the added noise. 

In our method, the training samples $x^{h}$ are drown from a HR video dataset. 
We use ControlNet \cite{controlnet} as an Encoder $E$ for LR videos to extract multi-scale latent features, which are injected into the feature maps of the denoising UNet $D$. In this way, $D$ is conditioned on LR videos to reconstruct the corresponding HR videos. Given the noise target $\epsilon_t$, the objective function for training $D$ and $E$ is as follows:
\begin{equation}\label{diffusion_loss} 
    \vspace{-0.1cm}
    \mathop{\mathbb{E}}_{t,x^h}||D(x^h_t, t,E(x^l))-\epsilon_t||^2,
    \vspace{-0.2cm}
\end{equation}
The stochastic nature of the diffusion denoising process leads to temporal instability in extended video sequences for VSR tasks.
Two essential problems need to be addressed simultaneously: modeling the spatial-temporal dependencies between adjacent frames, and extracting clean and stable features for accurate reconstruction in the presence of unknown and complex video degradations.
As depicted in Figure \ref{fig:framework}, our framework integrates the Spatial-Temporal Continuous Scan with the STCM into the Latent Diffusion Model (LDM)~\cite{rombach2022high} to ensure spatial-temporal coherence both within and across frame segments. Additionally, we introduce an advanced self-supervised learning approach, the Self-supervised Controlnet (MoCoCtrl), aimed at utilizing HR and LR video pairs to effectively refine the model's ability to generate detailed and noise-robust representations for VSR.

\subsection{Mamba for 3D Attention}
We provide a brief overview of SSM and present how to introduce it for real-world VSR.
\subsubsection{Preliminaries: State Space Models}
Drawing from the Kalman filter \cite{kalman1960new}, SSMs can be treated as linear time-invariant (LTI) systems that transform the input signal $x(t) \in \mathbb{R}$ to output signal $y(t) \in \mathbb{R}$ via the hidden state $\mathbf{h}(t) \in \mathbb{R}^{N}$. In essence, continuous-time SSMs can be represented as linear ordinary differential equations (ODEs), where they encode and decode one-dimensional sequential inputs. 
\begin{equation}\label{eq:ssm}
\begin{split}
  \mathbf{h'}(t) &= \mathbf{A} \mathbf{h}(t) + \mathbf{B} x(t),  \\
  y(t)  &= \mathbf{C} \mathbf{h}(t) + D x(t),
  \vspace{-0.1cm}
\end{split}
\end{equation}
where $\mathbf{A} \in \mathbb{R}^{N\times N}$, $\mathbf{B}\in \mathbb{R}^{N\times 1}$, $\mathbf{C}\in \mathbb{R}^{1\times N}$, and $D \in \mathbb{R}^{1}$ are the weighting parameters.

Usually, natural language and two-dimensional vision inputs are discrete signals. Therefore, Mamba utilizes the zero-order hold (ZOH) rule for discretization. Consequently, the ODEs can be iteratively resolved:

\begin{equation}
\begin{aligned}
    &\mathbf{\bar{A}} = e^{ \mathbf{\Delta} \mathbf{A} }, \\
    &\mathbf{\bar{B}} = \left( \mathbf{\Delta} \mathbf{A} \right)^{-1} \left( e^{  \mathbf{\Delta} \mathbf{A} } -\mathbf{I} \right) \cdot \mathbf{\Delta} \mathbf{B},\\
    &\mathbf{h}(t) = \mathbf{\bar{A}}~\mathbf{h}(t-1) + \mathbf{\bar{B}}~x(t),
\end{aligned}
\vspace{-0.1cm}
\end{equation}
Here, $\mathbf{\Delta}$ represents a model parameter. Mamba~\cite{mamba} aims to enhance the adaptability of SSMs by transitioning the time-invariant parameters into time-varying ones. This adjustment entails substituting the fixed model weights $(\mathbf{B}, \mathbf{C}, \mathbf{\Delta})$ with dynamic weights~\cite{dynamic_filter} that are dependent on the input $x$. This procedure, involving input-dependent parameters, is referred to as selective scanning.

\subsubsection{Spatial-Temporal Continuous Mamba}
Joint spatial-temporal modeling of videos plays a pivotal role in VSR. Approaches like Upscale-A-Video \cite{upscale} incorporate 3D convolution layers and temporal attention, yet their receptive field is constrained. On the other hand, recent video generation methods \cite{pku_yuan_lab_and_tuzhan_ai_etc_2024_10948109,polyak2024movie,bao2024vidu} leverage full 3D attention for motion modeling, demonstrating superior performance against decoupled spatial and temporal attention, however at the expense of a much higher computation complexity. In this study, we introduce Spatial-Temporal Continuous Mamba (STCM), to strike a balance between efficiency and effectiveness. \\
\textbf{3D-Mamba Block.} 
The 3D-Mamba Block, shown in Figure \ref{fig:framework}, is the core component of the STCM framework, specifically adapted for video-based tasks. It enhances spatiotemporal feature extraction by employing 3D depth-wise convolutions that capture both spatial and temporal dependencies. The block processes the input feature map using K types of scanning operations, generating sequences that are efficiently processed by a State Space Model (SSM)~\cite{mamba} to capture global context with linear complexity. After processing through the SSM, the K sequences are combined, resulting in an output feature map that maintains the original input dimensions. Details of the scanning operations will be discussed later.\\
\begin{figure}
    \centering
    \includegraphics[width=0.85\linewidth]{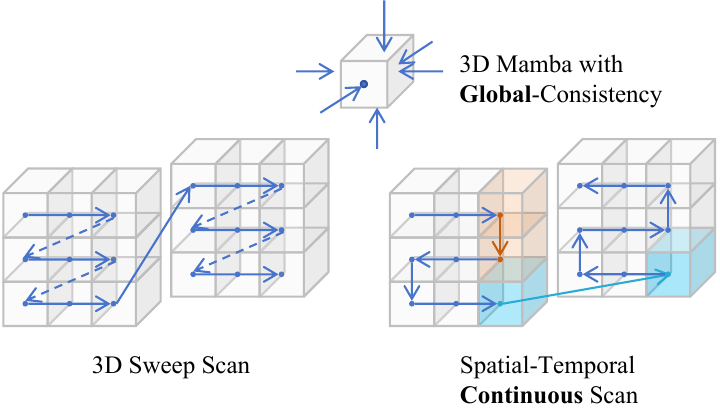}
    \vspace{-0.2cm}
    \caption{Diagram of Temporal-Spatial Continuous Scan Strategy with Global Consistency. Highlights a single continuous scan pathway across frames, emphasizing spatial-temporal alignment through intra-frame (\textcolor{orange}{orange}) and inter-frame (\textcolor{cyan}{cyan}) continuity.}
    \label{fig:scan}
    \vspace{-0.4cm}
\end{figure}
\textbf{Spatial-Temporal Continuous Scan.} 
As depicted in Figure \ref{fig:framework}, the 3D-Mamba Block utilizes the Spatial-Temporal Continuous Scan strategy to ensure a smooth, continuous information flow across both spatial and temporal dimensions. The scan involves three primary patterns, each featuring two distinct scanning trajectories: the original pattern and its flipped counterpart. This configuration results in a total of $K=6$ scanning paths.
Figure \ref{fig:scan} visually highlights a continuous scan pathway across frames, represented by the orange and cyan areas. The orange areas show intra-frame continuity, where the scan follows a sequential, pixel-by-pixel approach across the horizontal and vertical dimensions, capturing spatial information in a dense, continuous manner. The cyan areas indicate inter-frame continuity, where the scan tracks the same spatial points across successive frames, preserving pixel positions over time. Unlike traditional 3D sweep scan~\cite{xing2024segmamba}, which flattens the input and resets both within and between frames, our method maintains a continuous scan path, ensuring spatial-temporal coherence.
This approach is key to capturing consistent features in video data by maintaining pixel continuity both within and across frames, ensuring accurate temporal modeling. Our experiments in Section \ref{subsec:abla} validate its superiority over traditional methods.

\subsection{Momentum Contrastive ControlNet}
\label{Momentum Contrastive ControlNet}
The proposed 3D-Mamba module empowers the model with the capability to capture global spatiotemporal correlations. However, we find that 
directly training the model with Eq.~\ref{diffusion_loss} conditioning on LR videos leads to unstable training and emergence of artifacts. The presence of unknown and complex degradation in the LR videos causes optimization difficulty. To stabilize the training progress, we propose to provide additional supervising signal for the ControlNet with self-supervised learning leveraging the ground-truth HR videos.

\begin{figure}
    \centering
    \includegraphics[width=1\linewidth]{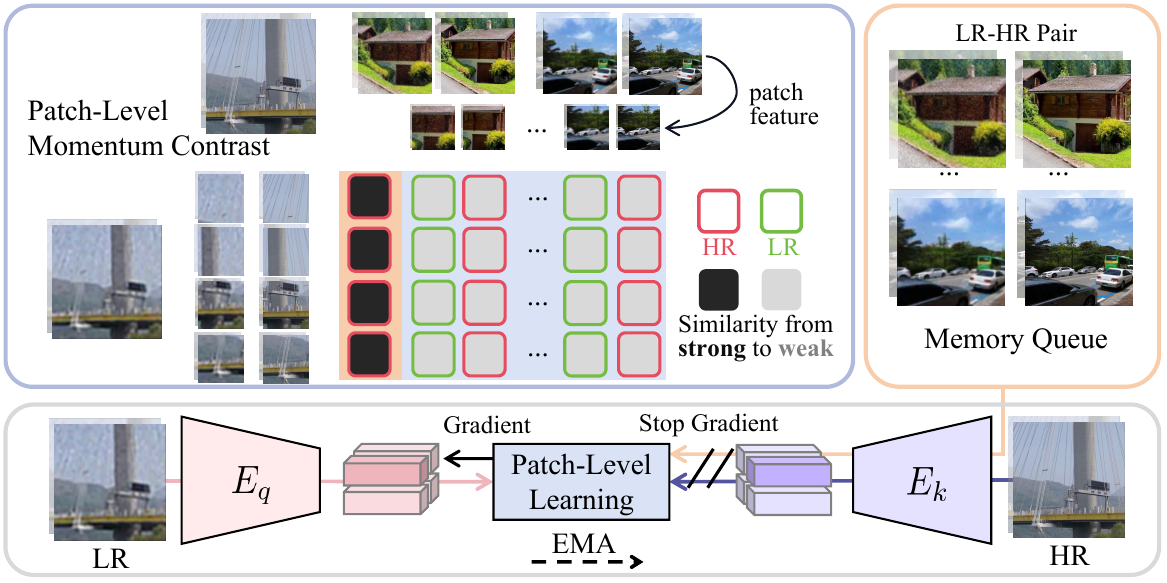}
    \vspace{-0.6cm}
    \caption{Patch-Level Momentum Contrast. LR and HR images are separately processed by ControlNet to extract feature maps, followed by patch-level contrastive learning on these features. The ControlNet for LR is online updated, whereas the ControlNet for HR updates its weights using a momentum way.}
    \label{fig:cl}
    \vspace{-0.3cm}
\end{figure}

As shown in Figure \ref{fig:cl}, we devise a MoCo-like \cite{self-supervised_task6} training framework in the optimization of the ControlNet, named MoCoCtrl. 
We choose the MoCo \cite{self-supervised_task6} framework due to its effectiveness and memory-friendly nature. While MoCo is traditionally applied to classification tasks with globally pooled features, our method adapts it for super-resolution tasks by focusing on patch-level features to capture finer spatial details.
In our MoCoCtrl framework, two encoders are used: a query encoder $E_q$ for LR frames and a momentum encoder $E_k$ for HR frames. The momentum encoder’s weights are updated as an exponential moving average (EMA) of the query encoder’s weights, ensuring stable representations of HR patches. During training, the positive sample pair $(x_{i}^l, x_{i}^{h})$ is mapped into feature maps through the query and key encoders as $q=E_q(x_{i}^{l})$ and $k_+=E_k(x_{i}^{h})$. To capture spatial details, we use a projection head to generate $P \times P$ patch features for each encoded feature map. Negative samples are selected from a memory queue, which is defined as $Q=\{E_k(x^h_j),(E_k(x_j^l)\mid j \in \{0, 1, \ldots, K/2\}, j \neq i\},$ where $K$ is the size of the memory queue. Both HR and LR samples are encoded in $Q$ to strengthen the contrastive learning signal and handle more diverse patch-level distinctions. 

For each encoded query $q$, a patch-level contrastive loss is defined as:
\begin{equation}
\mathcal{L}_q = \frac{1}{P^2} \sum_p -\log \frac{\exp(q^p \cdot k^p_+ / \tau)}{\exp(q^p \cdot k^p_+ / \tau) + \sum_{Q} \exp(q^p \cdot Q / \tau)},
\label{eq:infonce}
\end{equation}
where $q^p$ and $k^p_+$ represent the $p^{th}$ patch features of the query and positive HR sample. This patch-level contrastive approach improves the model’s ability to capture detailed spatial information, enhancing the performance of super-resolution tasks by precisely aligning LR and HR patches.
\begin{figure}
    \centering
    \includegraphics[width=1\linewidth]{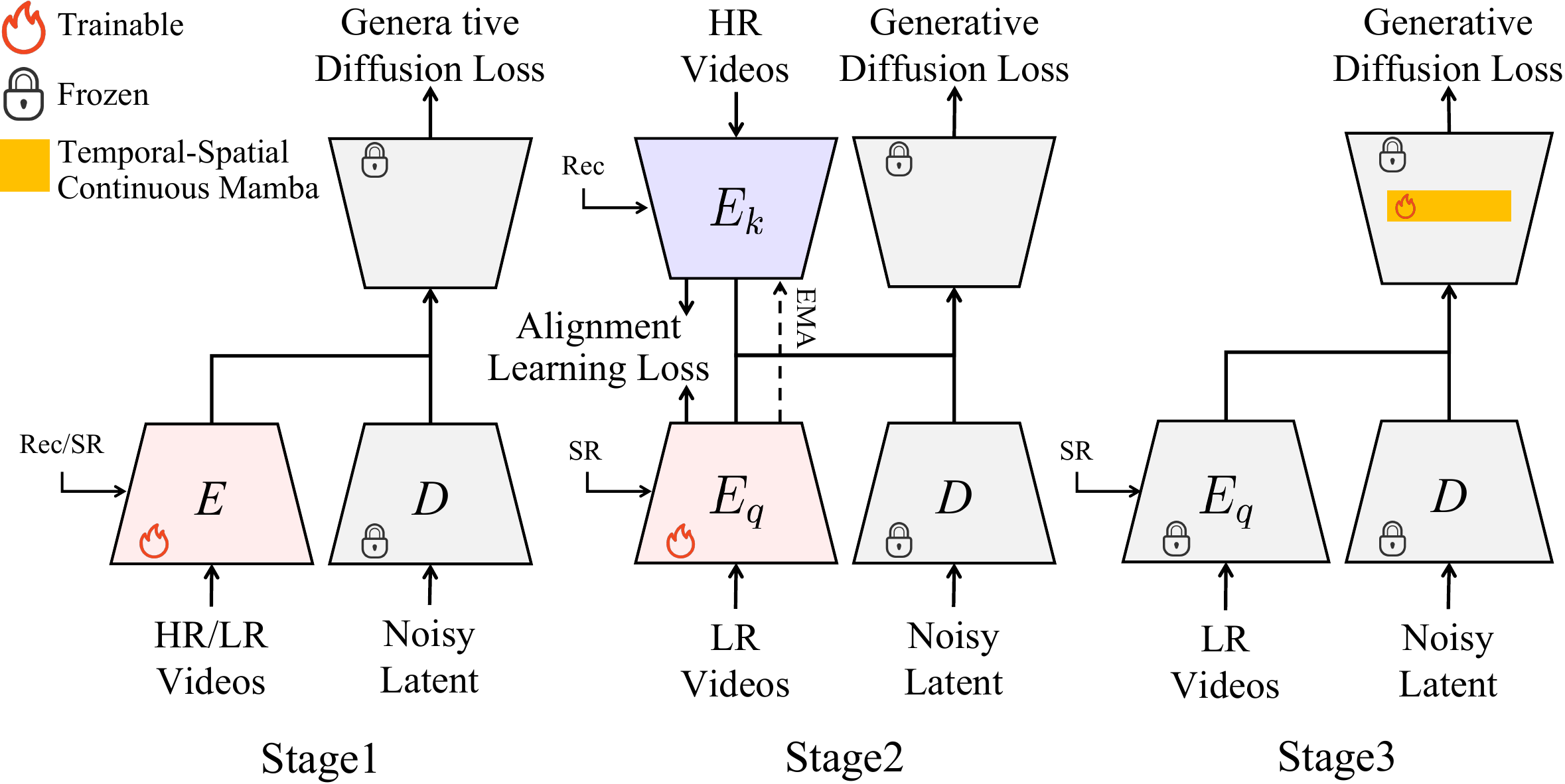}
    \vspace{-0.7cm}
    \caption{Multi-stage HR-LR hybrid training strategy. }
    \vspace{-0.5cm}
    \label{fig:3stage}
\end{figure}
\vspace{-0.4cm}
\subsection{Multi-stage HR-LR hybrid training strategy}
The proposed self-supervised ControlNet enables the model to find degradation-robust features, while the Mamba blocks model a comprehensive spatial-temporal correlation of the video sequences. We further propose a multi-stage HR-LR hybrid training strategy to facilitate the learning of the two modules.

We initialize the network using pretrained weights from Stable Diffusion V2.1. The weights of the original 2D U-Net are kept fixed, only the newly introduced layers are trained. As shown in Figure~\ref{fig:3stage}, the training consists of three stages. 
\textbf{In stage 1}, the ControlNet is trained with a mixture of HR/LR videos, where the HR videos can be viewed as LR videos with minimal degradations. When the inputs to the ControlNet are HR videos, the model is essentially trained to reconstruct the inputs. 
Training with HR videos allows the ControlNet to extract the most accurate features for reconstructing the HR videos. The mixture ratio of HR/LR videos starts at 1 and gradually decreases to 0.3, thereby gradually adapting the ControlNet for real-world VSR.
Additionally, a Reconstruction/SR label is introduced to enable the model to distinguish between the reconstruction and super-resolution tasks. 
\textbf{In stage 2}, the proposed MoCoCtrl is introduced. The training proceeds with a mixture of HR/LR videos, maintaining a fixed ratio at 1:1. 
The self-supervised learning more fully utilizes the reconstruction prior learned in Stage 1 to facilitate the training of real-world VSR.
\textbf{In stage 3}, the proposed Spatial-Temporal Continuous Mamba is integrated into the Unet. During this stage, the ControlNet remains unchanged. Only LR videos are used for training at this stage.

%% file: sec/4_experiments.tex
\vspace{-0.1cm}
\section{Experiments}
\label{sec:experiments}
\begin{table*}[!]
  \centering
  \footnotesize
  \renewcommand{\arraystretch}{0.4}

  \caption{\textbf{Quantitative comparisons of state-of-the-art VSR models on different VSR datasets.} The best and second performances are highlighted in \textcolor{red}{red} and \textcolor{blue}{blue}, respectively.}
\vspace{-0.2cm}
  \setlength{\tabcolsep}{1.5mm}{
  \begin{tabular}{l||l|c|c|c|c|c|c|c|c||c}
    \toprule[1.2pt]
    Datasets & Metrics & Bicubic  & RealESRGAN & StableSR & DBVSR  & RealBasicVSR  & RealViformer & Upscale-A-Video  & MGLD &\textbf{SCST} \\
    \midrule
    \multirow{4}{*}{REDS4} 
    & PSNR $\uparrow$ & 23.81 & 22.69 & 22.64 & 22.38 &  \textcolor{blue}{23.94}& \textcolor{red}{24.28} & 22.73 & 22.56 & 23.02\\
    & SSIM $\uparrow$ & 0.6313 &0.6201 & 0.6256 & 0.6015 & \textcolor{red}{0.6534}&  \textcolor{blue}{0.6513 }& 0.5982 & 0.5943 & 0.6108\\
    & LPIPS $\downarrow$ & 0.6485 &0.2964 & 0.2992 & 0.4941 & 0.2545 &\textcolor{blue}{ 0.2536 } & 0.3639 & 0.2660 & \textcolor{red}{0.2518}\\
    & DISTS $\downarrow$ & 0.2858 &0.1426 & 0.1277 & 0.2510 & 0.1196&  0.1306&  0.1840 &\textcolor{blue}{0.1171 }&\textcolor{red}{0.1094}\\
    \midrule
    \multirow{4}{*}{UDM10} 
    & PSNR $\uparrow$ & 26.40 &25.66 & 25.54 & 24.88 &26.10 &  \textcolor{red}{27.18}& 25.65 & 25.89 &\textcolor{blue}{26.42} \\
    & SSIM  $\uparrow$ & 0.7727 & 0.7817&  0.7622 & 0.7343& 0.7658&\textcolor{red}{0.7948}& 0.7413 & 0.7713 &\textcolor{blue}{0.7893}\\
    & LPIPS $\downarrow$ & 0.5039 &0.2739 & 0.2567 & 0.4685& 0.2812&  \textcolor{blue}{0.2580}& 0.2799 & 0.2551 & \textcolor{red}{0.2156}\\
    & DISTS $\downarrow$ & 0.2632 &0.1595 &\textcolor{blue}{ 0.1343} & 0.2454 & 0.1619&  0.1533& 0.1544 & 0.1386 & \textcolor{red}{0.1328}\\
    \midrule
    \multirow{4}{*}{SPMCS} 
    & PSNR $\uparrow$ & \textcolor{blue}{23.21} &22.38 & 22.21 & 22.01 & 23.10&\textcolor{red}{23.43 } & 21.72  & 22.87 & 22.17 \\
    & SSIM  $\uparrow$ & 0.6082 &0.6029 & 0.5932 & 0.5650 & 0.6049 & \textcolor{red}{0.6215}& 0.5327  & 0.6095 & \textcolor{blue}{0.6098} \\
    & LPIPS $\downarrow$ & 0.6360 &0.3238 & 0.3079 & 0.5160 &0.3142 & \textcolor{blue}{0.3030} &0.3743  & 0.3041 & \textcolor{red}{0.2600} \\
    & DISTS $\downarrow$ & 0.3092 & 0.1924& \textcolor{blue}{0.1709}& 0.2725 & 0.1847&  0.1884&0.2201 & 0.1769 & \textcolor{red}{0.1612} \\
    \midrule
    \multirow{4}{*}{YouHQ40} 
    & PSNR $\uparrow$ & \textcolor{red}{24.47} &23.76 & 23.41& 23.41 &23.26 &   \textcolor{blue}{24.44} &23.41  & 23.48 & 24.31\\
    & SSIM  $\uparrow$ & \textcolor{red}{0.6787} & 0.6743 & 0.6555 & 0.6480&  0.6306& 0.6730& 0.6252&0.6394 &\textcolor{blue}{0.6759}\\
    & LPIPS $\downarrow$ & 0.5437 &0.3126 &  \textcolor{blue}{0.3071} &0.4699 & 0.3706&0.3202  &0.3349 &0.3309 & \textcolor{red}{0.2525}\\
    & DISTS $\downarrow$ & 0.2416 &0.1537 & \textcolor{blue}{0.1360} &0.2137 & 0.1745&  0.1740& 0.1618& 0.1540& \textcolor{red}{0.1344}\\
    \midrule
    \multirow{5}{*}{VideoLQ} 
    & CLIP-IQA $\uparrow$ &0.2949 & 0.3617& \textcolor{blue}{0.4160} &0.2475 &0.3881& 0.3460& 0.2818 & 0.3462& \textcolor{red}{0.4859}\\
    & MUSIQ  $\uparrow$ & 22.56&49.84 &47.77 & 31.27& \textcolor{blue}{55.61} &52.09 & 43.34 &50.94 & \textcolor{red}{59.20}\\
    & NIQE $\downarrow$ & 8.059 &4.203 & 4.418& 6.278& \textcolor{blue}{3.698} & 4.057& 4.8762&3.727&\textcolor{red}{3.566} \\
    & DOVER $\uparrow$ &0.3882 &0.7152 &0.7029 & 0.5264& \textcolor{blue}{0.7367} &0.7194 &0.6199 &0.7340 &\textcolor{red}{0.7443} \\
    \bottomrule[1.2pt]
  \end{tabular}
  }
  \label{tab:exp_quan}
  \vspace{-0.2cm}
\end{table*}

\begin{figure*}
    \centering
    \includegraphics[width=1\linewidth]{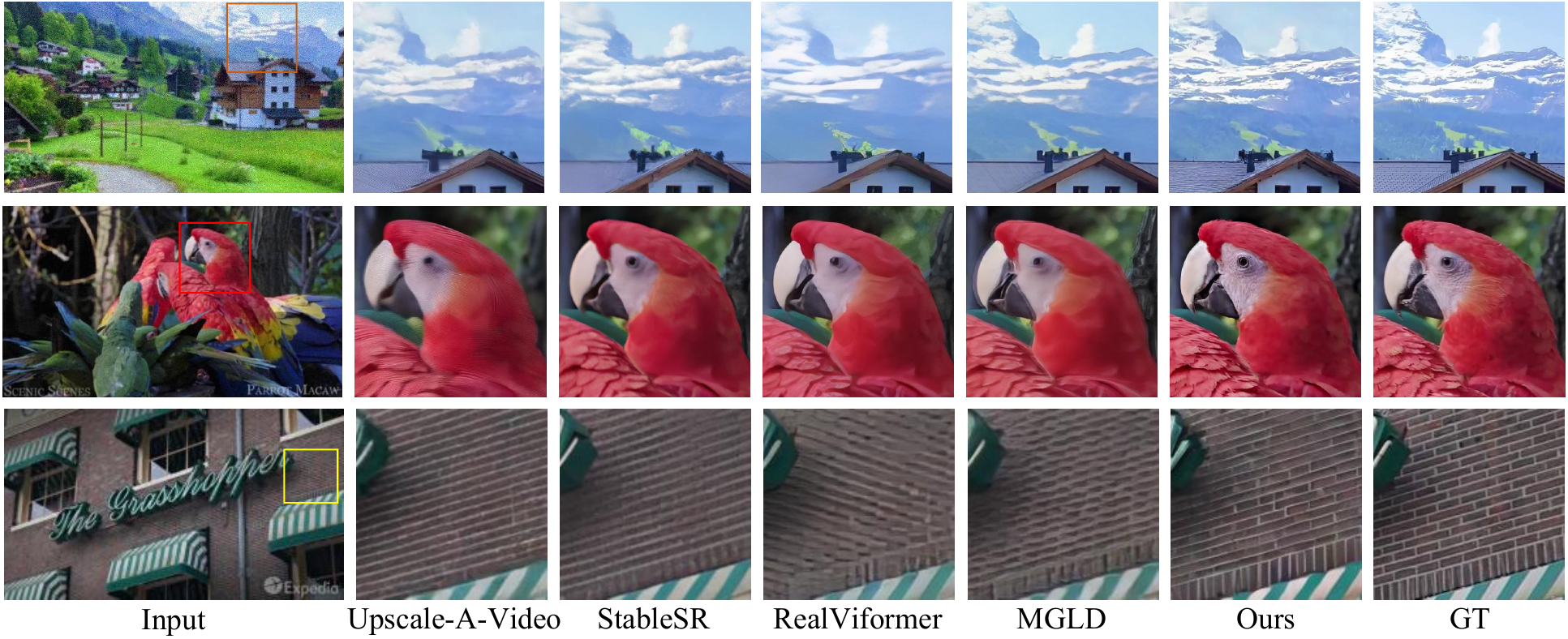}
    \vspace{-0.6cm}
    \caption{Qualitative comparisons on synthetic low-quality videos. \textbf{(Zoom-in for best view)}}
    \label{fig:youhq}
    \vspace{-0.5cm}
\end{figure*}
\subsection{Experimental settings}
\textbf{Training Datasets.} We train our model using REDS~\cite{nah2019ntire} and YouHQ~\cite{upscale} datasets. Following~\citet{wang2019edvr}, the REDS4 dataset\footnote{Clips 000, 011, 015, 020 of REDS training set.} within REDS is excluded from training the model. Additionally, we split YouHQ40 dataset 
for testing only following~\citet{upscale}, which contains 40 videos. 
Following the degradation pipeline of RealBasicVSR~\cite{chan2022investigating}, we generate the LQ-HQ video pairs for training.\\
\textbf{Testing Datasets.} We construct the test set with four synthetic testing datasets (i.e., REDS4, UDM10 \cite{yi2019progressive}, SPMCS \cite{tao2017detail}, and YouHQ40), which follow the same degradation pipeline in training to generate LQ videos. Additionally, we evaluate the models on a real-world dataset VideoLQ~\cite{chan2022investigating}.\\
\textbf{Implementation Details.}
The model is trained on 8 NVIDIA A100 GPUs with Adam~\cite{kinga2015method} optimizer and a batch size of 32. The sequence length and video resolution are set to 8 and $512 \times 512$ respectively. To better leverage the prior knowledge of text-to-image diffusion models, we use Panda-70M model~\cite{chen2024panda70m} to generate text prompts during training and inference. Training takes approximately 12 hours for stage 1, 30 hours for stage 2, and 30 hours for stage 3.
During inference, owing to memory limitations, we segment LR videos into multiple sequences. The number of sampling steps is set to 20.\\
\textbf{Evaluation Metrics.}
In order to comprehensively evaluate real-world VSR methods, we utilize a range of metrics across both synthetic and real-world datasets.
For synthetic datasets, we assess the video quality using four prevalent metrics in real-world VSR tasks: learned perceptual image patch similarity (LPIPS) \cite{LPIPS}, deep image structure and texture similarity (DISTS) \cite{DISTS}, structural similarity index (SSIM), and the widely recognized peak signal-to-noiseratio (PSNR). 
When assessing the performance on the real-world dataset, we compute the
no-reference image quality metrics: the natural image quality evaluator (NIQE) \cite{NIQE} and CLIP-IQA \cite{wang2023exploring}.
Additionally, we also include a deep learning based image-based metric MUSIQ \cite{MUSIQ} and a video quality assessment metric DOVER \cite{DOVER}.
\begin{figure*}
    \centering
    \includegraphics[width=1\linewidth]{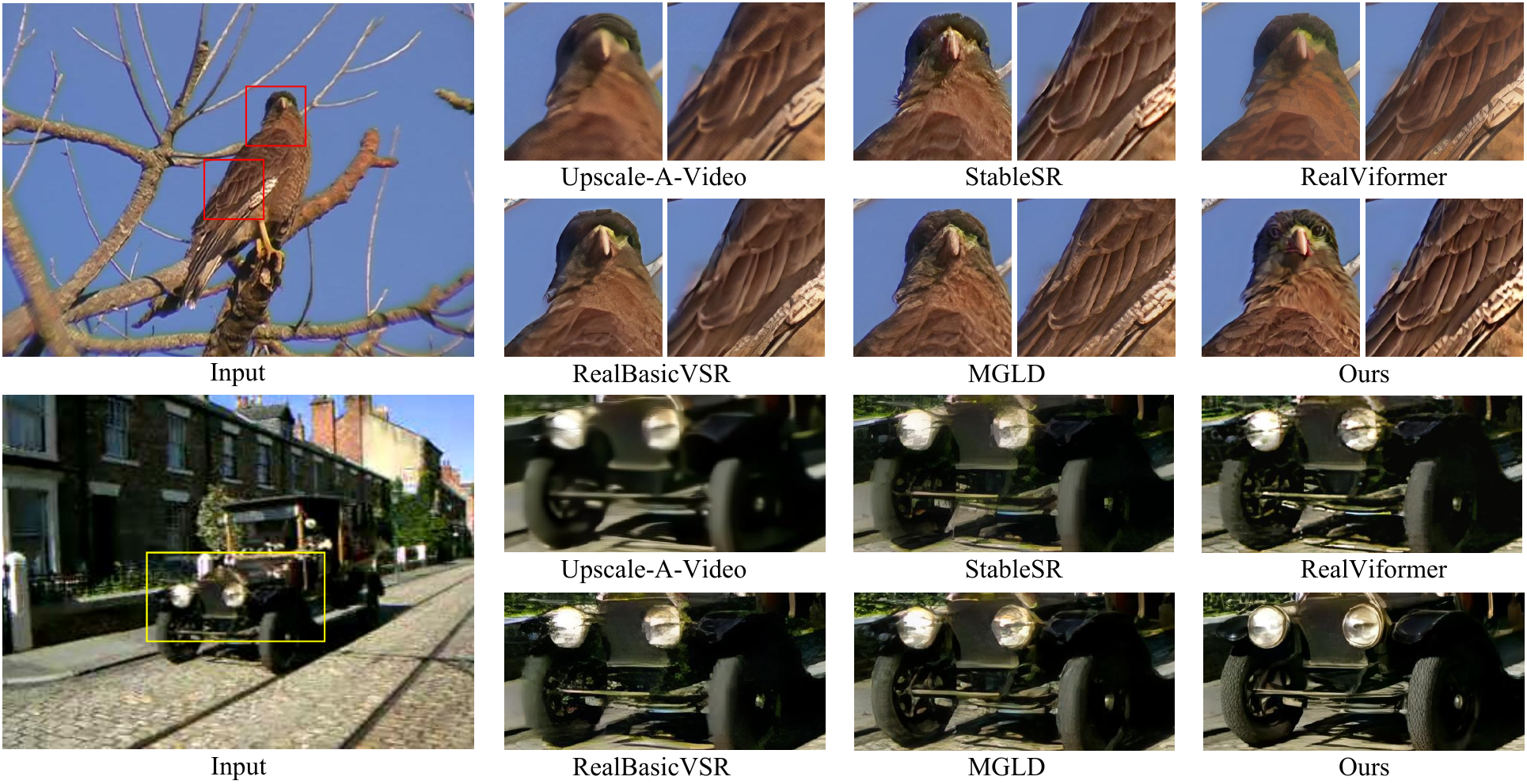}
    \vspace{-0.6cm}
    \caption{Qualitative comparisons on real-world test videos in VideoLQ. \textbf{(Zoom-in for best view)}}
    \label{fig:videolq}
    \vspace{-0.5cm}
\end{figure*}
\subsection{Experimental Results}
To comprehensively evaluate the performance of our SCST algorithm, we compare it with several state-of-the-art methods, including two real-world image super-resolution models (RealESRGAN \cite{RealESRGAN}, StableSR \cite{stablesr}), two real-world VSR models (DBVSR \cite{DBVSR}, RealBasicVSR \cite{realbasicvsr}, and recently proposed Upscale-A-Video~\cite{upscale}, MGLD~\cite{mgld} and RealViformer~\cite{zhang2024realviformer}. \\
\textbf{Quantitative Comparison.}
Table \ref{tab:exp_quan} demonstrates the quantitative comparison on the synthetic datasets and real-world video benchmarks.
From the Table \ref{tab:exp_quan}, we can observe that our approach attains superior performance in terms of full-reference perceptual metrics LPIPS and DISTS across all synthetic test datasets. This suggests that our method can effectively restore high-quality, realistic details from sequences affected by intricate degradations.
While methods such as DBVSR might exhibit improved performance concerning PSNR or SSIM on specific datasets, they often produce blurred outputs, as evidenced by the LPIPS and DISTS metrics.
When considering the performance on the real-world VSR dataset VideoLQ, our approach achieves the best results in CLIP-IQA, MUSIQ , NIQE, and DOVER metrics, which indicate the robust capacity of our SCST to enhance real-world videos, producing authentic details and clean textures.\\
\textbf{Qualitative Comparison.}
To further demonstrate the effectiveness of our SCST, we conduct visual comparisons of these models on both synthetic datasets and real-world VideoLQ dataset, as shown in Figure \ref{fig:youhq} and Figure \ref{fig:videolq}, respectively.
For synthetic datasets, from the Figure \ref{fig:youhq}, we can observe that SCST excels in reconstructing structures while generating cleaner details under complex degradations. The improvements are particularly evident in the enhanced clarity of distant mountain peaks and the textured details of brick walls, as well as the naturalistic rendering of a parrot's plumage.
For the real-world VSR, it is apparent that SCST surpasses other state-of-the-art algorithms in eliminating intricate spatially varying degradations while producing realistic details. Note that, SCST excels as the sole method capable of accurately delineating the intricate details of the eagle's eyes, showcasing its advanced resolution capabilities. Similarly, SCST provides a significantly clearer depiction of the vehicle's tires, accurately capturing the texture and contours with high fidelity. In contrast, other state-of-the-art methods result in blurred and less defined features.
\vspace{-0.5cm}
\subsection{Ablation Study}
\begin{figure}
    \centering
    \includegraphics[width=1\linewidth]{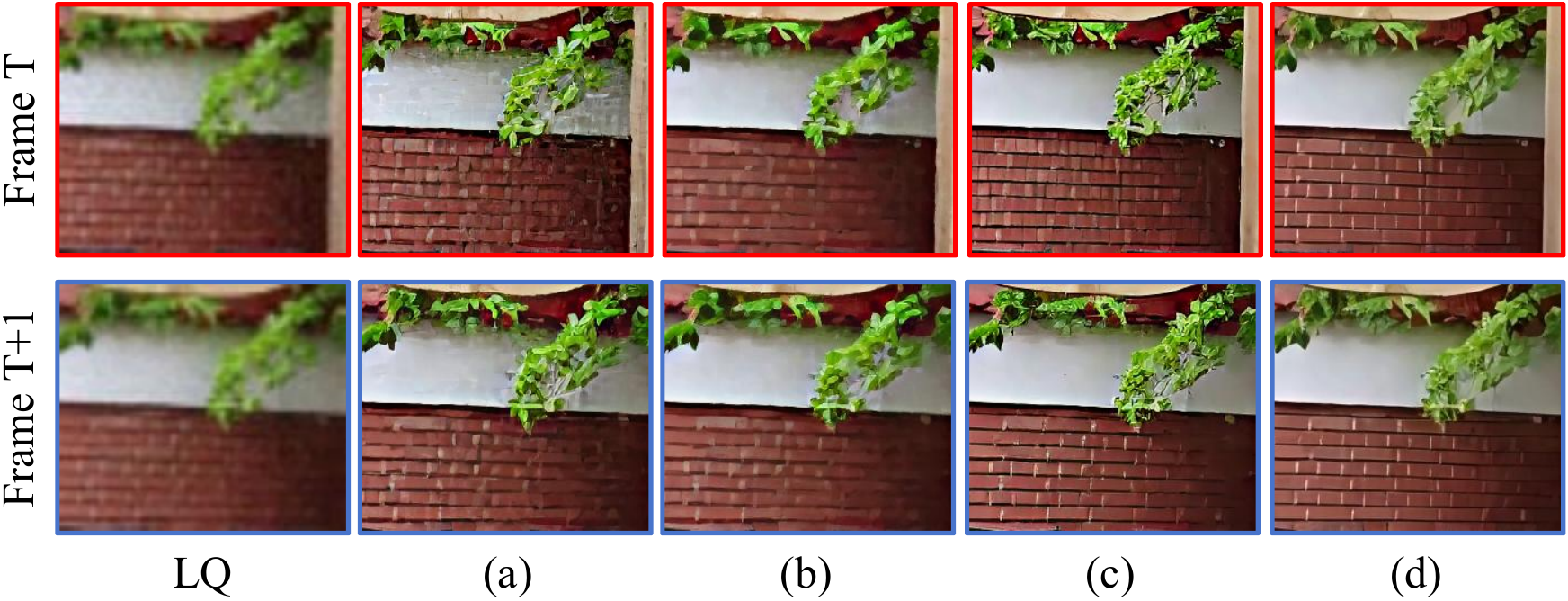}
    \vspace{-0.7cm}
    \caption{Comparison of two adjacent frames in one video SR results synthesized by different components.}
    \vspace{-0.3cm}
    \label{fig:ablation}
\end{figure}

\begin{figure}
    \centering
    \includegraphics[width=\linewidth]{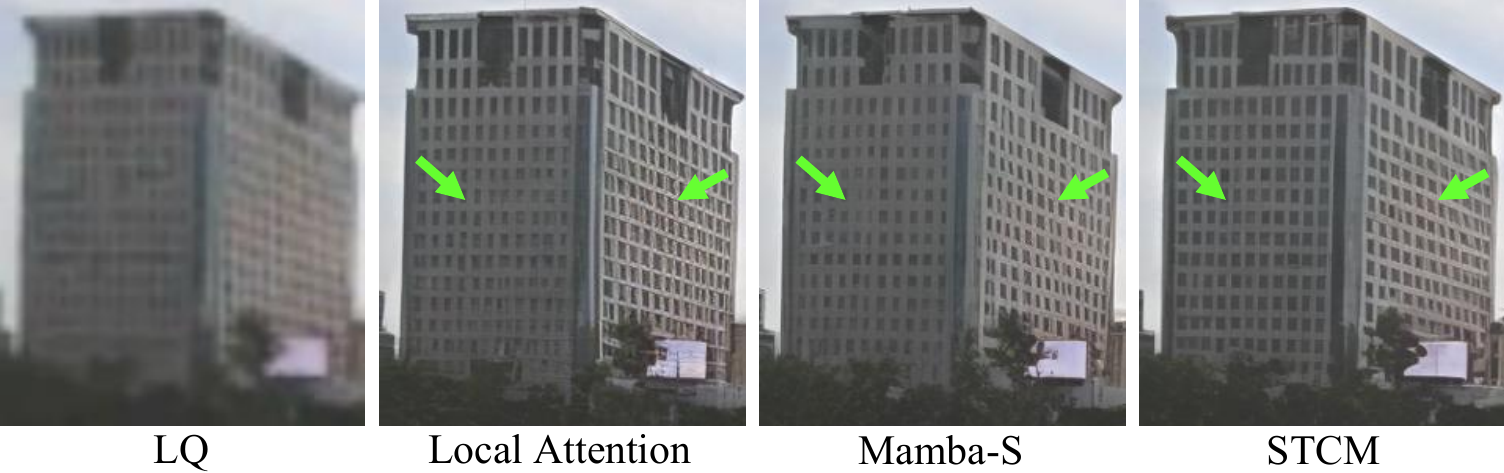}
    \vspace{-20pt}
    \caption{Different Spatial-temporal modeling approaches}
    \vspace{-0.5cm}
    \label{fig:mamba2}
\end{figure}

\begin{table}[t]
	\caption{Ablation study on the different components of SCST on YouHQ. Best marked in \textbf{bold}.}
    \vspace{-0.15cm}
	\centering
	\label{tab:ablation}
	\resizebox{0.45\textwidth}{!}{
	\begin{tabular}{c|cc|cccc}
    \toprule[1.2pt]
        Models & MoCoCtrl & STCM & PSNR $\uparrow$ & SSIM $\uparrow$ & LPIPS $\downarrow$ & DISTS $\downarrow$ \\ 
        \hline
        (a) &  &  & 21.22 &0.6357&0.2824&0.1596 \\
        (b) &  & \checkmark &23.63 &0.6563& 0.2671&0.1473  \\ 
        (c) &\checkmark & &23.18 & 0.6533&0.2581&0.1470\\
        (d) & \checkmark & \checkmark & \textbf{24.31} & \textbf{0.6758} & \textbf{0.2525} & \textbf{0.1344}\\ 
    \toprule[1.2pt]
        \end{tabular}}
     \vspace{-0.5cm}
\end{table}
\label{subsec:abla}
\textbf{Baseline Design.} To assess the impact of the two key components of our network, MoCoCtrl and STCM, we conducted a series of ablation experiments. We start by creating a baseline model removing all the major components of the network. Specifically, the baseline model excludes the second-stage contrastive training and directly trains the model using Eq.~\ref{diffusion_loss}. \\
\textbf{Effectiveness of MoCoCtrl.}
As shown in Figure \ref{fig:ablation} (c) and (d), as well as Table \ref{tab:ablation}, the MoCoCtrl module results in clearer super-resolution outputs and better performance metrics. We can observe that Model (d) outperforms Model (b) in terms of PSNR, SSIM, and LPIPS. Specifically, Model (d) shows better performance in PSNR (24.31 dB vs. 23.63 dB), higher SSIM (0.6758 vs. 0.6563), lower LPIPS (0.2525 vs. 0.2671) and DISTS (0.1344 vs. 0.1473). This comparison clearly demonstrates the improvement in video super-resolution quality in Model (d) due to the MoCoCtrl module, thus confirming its effectiveness. 
\\
\textbf{Effectiveness of STCM.} In addition to the proposed MoCoCtrl, STCM further enhances the quality of our generated videos. Specifically, STCM extends its global receptive field to capture complex spatio-temporal relationships and employs continuous scanning to identify  long-range dependencies within video sequences. 
STCM is designed to facilitate the model in better understanding video content, thereby enhancing temporal consistency while maintaining high-resolution output with increased fidelity. As illustrated in Figure~\ref{fig:ablation}, the temporal consistency is markedly inferior in the absence of the STCM module.
Moreover, the integration of the STCM module into Model (c) is shown to yield substantial enhancements in performance metrics, with noteworthy increases of 1.13 dB in PSNR, 0.0225 in SSIM, 0.0056 in LPIPS, and 0.0126 in DISTS, as demonstrated in Table~\ref{tab:ablation}.\\
\textbf{Analysis on the Mamba design.}
\begin{table}[t]
\centering
\caption{Comparison of different spatial-temporal modeling approaches on YouHQ and REDS4. “Mamba-S” employs a 3D Sweep Scan strategy. Best marked in \textbf{blod}.} 
\vspace{-6pt}
	\centering
	\label{tab:stcm}
	\resizebox{0.47\textwidth}{!}{
    \begin{tabular}{c | c c | c c}
    \toprule[1.2pt]
        \multirow{2}{*}{Models} & \multicolumn{2}{c|}{PSNR$\uparrow$ / LPIPS$\downarrow$} & \multicolumn{2}{c}{Warp Error$\downarrow$} \\ 
        \cline{2-5}
        & YouHQ & REDS4 & YouHQ & REDS4 \\
        \hline
        w/o Temporal & 23.18 / 0.2581 & 
        22.73 / 0.2669 & 0.3603 & 3.619 \\
        Local Attention & 24.07 / 0.2570  &  22.60 / 0.2629 & 0.2311 & 2.889 \\
        Mamba-S  & 24.12 / 0.2533  &  22.59 / 0.2590 & 0.2494 & 2.926\\ 
        STCM  & \textbf{24.31} / \textbf{0.2525}  &  \textbf{23.02} / \textbf{0.2518} & \textbf{0.2295} & \textbf{2.878} \\ 
    \bottomrule[1.2pt]
    \end{tabular}}
    \vspace{-0.5cm}
\end{table}
To further elucidate the advantages of our STCM, we
also conducted an analysis comparing various spatial-temporal
modeling approaches in terms of restoration quality and temporal consistency. We replace the STCM component with local inter-frame attention (Local Attention) \cite{guo2023animatediff} and 3D Sweep Scan Mamba (Mamba-S) to compare different spatial-temporal modeling approaches. Table~\ref{tab:stcm} shows STCM achieves the best performance among all methods in PSNR and LPIPS.
Compared to Local Attention, which is limited by its spatial receptive field, STCM leverages \emph{multi-direction fusion} to fully capture the 3D data pattern. As shown in Figure~\ref{fig:mamba2}, Local Attention produces visible distortions and blurred edges, especially in complex regions, resulting in misaligned geometry. In contrast, STCM incorporates surrounding pixel information to reconstruct more rectilinear and well-aligned structures, significantly reducing artifacts in challenging areas.
\begin{figure}
    \centering
    \includegraphics[width=\linewidth]{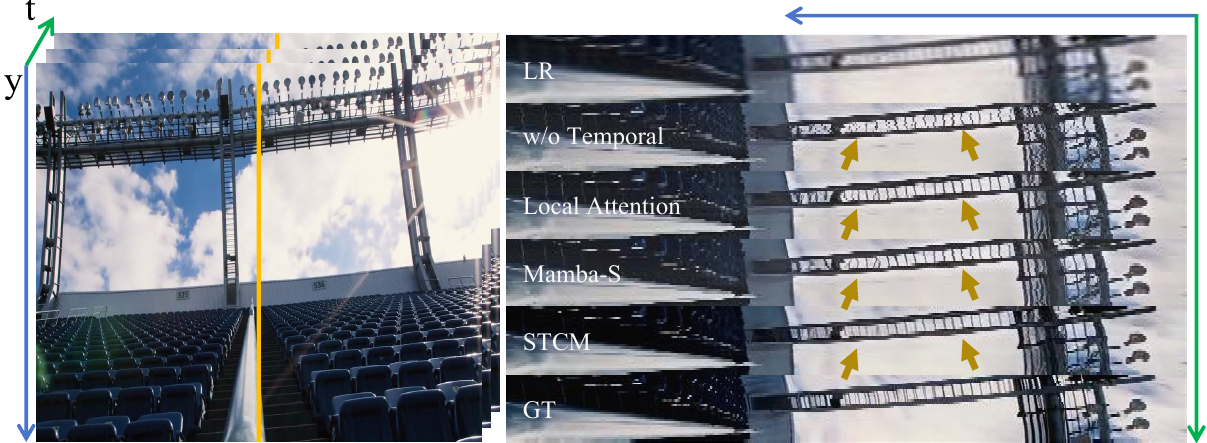}
    \vspace{-0.6cm}
    \caption{Visual comparison on temporal profile with different spatial-temporal modeling approaches, with STCM exhibiting the best temporal consistency. \textbf{(Zoom-in for best view)}}
    \label{fig:temporal}
    \vspace{-0.6cm}
\end{figure}
Beyond restoration quality, we use the Warping Error (WE) \cite{lai2018learningwe} to measure temporal consistency. Table \ref{tab:stcm} clearly shows that STCM outperforms all other methods on both datasets, achieving the lowest WE scores of 0.2295 and 2.878 on YouHQ and REDS4, respectively.
It is worth mentioning that the excellent performance of STCM is driven by its spatial-temporal continuous scanning strategy. Unlike Mamba-S, which employs a 3D Sweep Scan that disrupts continuity by resetting between frames, our approach maintains an uninterrupted flow of information. This continuous scanning ensures spatial-temporal coherence and precise frame alignment. Figure \ref{fig:temporal} illustrates that STCM preserves steady structures across frames, while Mamba-S exhibits temporal fluctuations and misalignments. 

%% file: sec/5_conclusion.tex
\section{Conclusion}
\label{sec:conclusion}
In this paper, we propose a Self-supervised ControlNet with Spatio-Temporal Mamba algorithm dubbed SCST, designed for high-quality and temporally consistent real-world Video Super-Resolution (VSR). The distinctiveness of our proposed method lies in the idea of introducing a specific Self-supervised ControlNet as a degradation removal module, reducing the impact of complex degradations on the effectiveness of VSR. To further model spatio-temporal relationships for temporal consistency, we present an efficient 3D attention-based variant of the successful Mamba model. Finally, to ensure training stability, we propose a multi-stage HR-LR hybrid training strategy, decomposing real-world VSR into multiple subtasks, with each stage addressing a specific task. Our proposed SCST has achieved state-of-the-art results on existing VSR benchmarks.

%% file: main.bbl
\begin{thebibliography}{75}
\providecommand{\natexlab}[1]{#1}
\providecommand{\url}[1]{\texttt{#1}}
\expandafter\ifx\csname urlstyle\endcsname\relax
  \providecommand{\doi}[1]{doi: #1}\else
  \providecommand{\doi}{doi: \begingroup \urlstyle{rm}\Url}\fi

\bibitem[Assran et~al.(2022)Assran, Caron, Misra, Bojanowski, Bordes, Vincent, Joulin, Rabbat, and Ballas]{cl1}
Mahmoud Assran, Mathilde Caron, Ishan Misra, Piotr Bojanowski, Florian Bordes, Pascal Vincent, Armand Joulin, Mike Rabbat, and Nicolas Ballas.
\newblock Masked siamese networks for label-efficient learning.
\newblock In \emph{European Conference on Computer Vision}, pages 456--473. Springer, 2022.

\bibitem[Bao et~al.(2024)Bao, Xiang, Yue, He, Zhu, Zheng, Zhao, Liu, Wang, and Zhu]{bao2024vidu}
Fan Bao, Chendong Xiang, Gang Yue, Guande He, Hongzhou Zhu, Kaiwen Zheng, Min Zhao, Shilong Liu, Yaole Wang, and Jun Zhu.
\newblock Vidu: a highly consistent, dynamic and skilled text-to-video generator with diffusion models.
\newblock \emph{arXiv preprint arXiv:2405.04233}, 2024.

\bibitem[Cao et~al.(2021)Cao, Li, Zhang, and Van~Gool]{cao2021video}
Jiezhang Cao, Yawei Li, Kai Zhang, and Luc Van~Gool.
\newblock Video super-resolution transformer.
\newblock \emph{arXiv preprint arXiv:2106.06847}, 2021.

\bibitem[Caron et~al.(2020)Caron, Misra, Mairal, Goyal, Bojanowski, and Joulin]{self-supervised_task1}
Mathilde Caron, Ishan Misra, Julien Mairal, Priya Goyal, Piotr Bojanowski, and Armand Joulin.
\newblock Unsupervised learning of visual features by contrasting cluster assignments.
\newblock \emph{Advances in neural information processing systems}, 33:\penalty0 9912--9924, 2020.

\bibitem[Caron et~al.(2021)Caron, Touvron, Misra, J{\'e}gou, Mairal, Bojanowski, and Joulin]{self-supervised_task2}
Mathilde Caron, Hugo Touvron, Ishan Misra, Herv{\'e} J{\'e}gou, Julien Mairal, Piotr Bojanowski, and Armand Joulin.
\newblock Emerging properties in self-supervised vision transformers.
\newblock In \emph{Proceedings of the IEEE/CVF international conference on computer vision}, pages 9650--9660, 2021.

\bibitem[Carrillo et~al.(2023)Carrillo, Cl{\'e}ment, Bugeau, and Simo-Serra]{carrillo2023diffusart}
Hernan Carrillo, Micha{\"e}l Cl{\'e}ment, Aur{\'e}lie Bugeau, and Edgar Simo-Serra.
\newblock Diffusart: Enhancing line art colorization with conditional diffusion models.
\newblock In \emph{Proceedings of the IEEE/CVF Conference on Computer Vision and Pattern Recognition}, pages 3486--3490, 2023.

\bibitem[Chan et~al.(2021)Chan, Wang, Yu, Dong, and Loy]{chan2021basicvsr}
Kelvin~CK Chan, Xintao Wang, Ke Yu, Chao Dong, and Chen~Change Loy.
\newblock Basicvsr: The search for essential components in video super-resolution and beyond.
\newblock In \emph{Proceedings of the IEEE/CVF conference on computer vision and pattern recognition}, pages 4947--4956, 2021.

\bibitem[Chan et~al.(2022{\natexlab{a}})Chan, Zhou, Xu, and Loy]{chan2022basicvsr++}
Kelvin~CK Chan, Shangchen Zhou, Xiangyu Xu, and Chen~Change Loy.
\newblock Basicvsr++: Improving video super-resolution with enhanced propagation and alignment.
\newblock In \emph{Proceedings of the IEEE/CVF conference on computer vision and pattern recognition}, pages 5972--5981, 2022{\natexlab{a}}.

\bibitem[Chan et~al.(2022{\natexlab{b}})Chan, Zhou, Xu, and Loy]{chan2022investigating}
Kelvin~CK Chan, Shangchen Zhou, Xiangyu Xu, and Chen~Change Loy.
\newblock Investigating tradeoffs in real-world video super-resolution.
\newblock In \emph{Proceedings of the IEEE/CVF Conference on Computer Vision and Pattern Recognition}, pages 5962--5971, 2022{\natexlab{b}}.

\bibitem[Chan et~al.(2022{\natexlab{c}})Chan, Zhou, Xu, and Loy]{realbasicvsr}
Kelvin~CK Chan, Shangchen Zhou, Xiangyu Xu, and Chen~Change Loy.
\newblock Investigating tradeoffs in real-world video super-resolution.
\newblock In \emph{Proceedings of the IEEE/CVF Conference on Computer Vision and Pattern Recognition}, pages 5962--5971, 2022{\natexlab{c}}.

\bibitem[Chen et~al.(2020)Chen, Kornblith, Norouzi, and Hinton]{self-supervised_task3}
Ting Chen, Simon Kornblith, Mohammad Norouzi, and Geoffrey Hinton.
\newblock A simple framework for contrastive learning of visual representations.
\newblock In \emph{International conference on machine learning}, pages 1597--1607. PMLR, 2020.

\bibitem[Chen et~al.(2024)Chen, Siarohin, Menapace, Deyneka, Chao, Jeon, Fang, Lee, Ren, Yang, et~al.]{chen2024panda70m}
Tsai-Shien Chen, Aliaksandr Siarohin, Willi Menapace, Ekaterina Deyneka, Hsiang-wei Chao, Byung~Eun Jeon, Yuwei Fang, Hsin-Ying Lee, Jian Ren, Ming-Hsuan Yang, et~al.
\newblock Panda-70m: Captioning 70m videos with multiple cross-modality teachers.
\newblock In \emph{Proceedings of the IEEE/CVF Conference on Computer Vision and Pattern Recognition}, pages 13320--13331, 2024.

\bibitem[Ding et~al.(2020)Ding, Ma, Wang, and Simoncelli]{DISTS}
Keyan Ding, Kede Ma, Shiqi Wang, and Eero~P Simoncelli.
\newblock Image quality assessment: Unifying structure and texture similarity.
\newblock \emph{IEEE transactions on pattern analysis and machine intelligence}, 44\penalty0 (5):\penalty0 2567--2581, 2020.

\bibitem[Feng and Patras(2022)]{cl2}
Chen Feng and Ioannis Patras.
\newblock Adaptive soft contrastive learning.
\newblock In \emph{2022 26th International Conference on Pattern Recognition (ICPR)}, pages 2721--2727. IEEE, 2022.

\bibitem[Feng and Patras(2023)]{cl3}
Chen Feng and Ioannis Patras.
\newblock Maskcon: Masked contrastive learning for coarse-labelled dataset.
\newblock In \emph{Proceedings of the IEEE/CVF Conference on Computer Vision and Pattern Recognition}, pages 19913--19922, 2023.

\bibitem[Feng et~al.(2021)Feng, Tzimiropoulos, and Patras]{self-supervised_task4}
Chen Feng, Georgios Tzimiropoulos, and Ioannis Patras.
\newblock Ssr: An efficient and robust framework for learning with unknown label noise.
\newblock \emph{arXiv preprint arXiv:2111.11288}, 2021.

\bibitem[Gao et~al.(2024)Gao, Feng, and Patras]{self-supervised_task5}
Zheng Gao, Chen Feng, and Ioannis Patras.
\newblock Self-supervised representation learning with cross-context learning between global and hypercolumn features.
\newblock In \emph{Proceedings of the IEEE/CVF Winter Conference on Applications of Computer Vision}, pages 1773--1783, 2024.

\bibitem[Gu and Dao(2023)]{mamba}
Albert Gu and Tri Dao.
\newblock Mamba: Linear-time sequence modeling with selective state spaces.
\newblock \emph{arXiv preprint arXiv:2312.00752}, 2023.

\bibitem[Gu et~al.(2021)Gu, Goel, and R{\'e}]{S4}
Albert Gu, Karan Goel, and Christopher R{\'e}.
\newblock Efficiently modeling long sequences with structured state spaces.
\newblock \emph{arXiv preprint arXiv:2111.00396}, 2021.

\bibitem[Guo et~al.(2023)Guo, Yang, Rao, Liang, Wang, Qiao, Agrawala, Lin, and Dai]{guo2023animatediff}
Yuwei Guo, Ceyuan Yang, Anyi Rao, Zhengyang Liang, Yaohui Wang, Yu Qiao, Maneesh Agrawala, Dahua Lin, and Bo Dai.
\newblock Animatediff: Animate your personalized text-to-image diffusion models without specific tuning.
\newblock \emph{arXiv preprint arXiv:2307.04725}, 2023.

\bibitem[Gupta et~al.(2022)Gupta, Gu, and Berant]{variant4}
Ankit Gupta, Albert Gu, and Jonathan Berant.
\newblock Diagonal state spaces are as effective as structured state spaces.
\newblock \emph{Advances in Neural Information Processing Systems}, 35:\penalty0 22982--22994, 2022.

\bibitem[He et~al.(2020)He, Fan, Wu, Xie, and Girshick]{self-supervised_task6}
Kaiming He, Haoqi Fan, Yuxin Wu, Saining Xie, and Ross Girshick.
\newblock Momentum contrast for unsupervised visual representation learning.
\newblock In \emph{Proceedings of the IEEE/CVF conference on computer vision and pattern recognition}, pages 9729--9738, 2020.

\bibitem[He et~al.(2022)He, Chen, Xie, Li, Doll{\'a}r, and Girshick]{self-supervised_task7}
Kaiming He, Xinlei Chen, Saining Xie, Yanghao Li, Piotr Doll{\'a}r, and Ross Girshick.
\newblock Masked autoencoders are scalable vision learners.
\newblock In \emph{Proceedings of the IEEE/CVF conference on computer vision and pattern recognition}, pages 16000--16009, 2022.

\bibitem[Ho et~al.(2020)Ho, Jain, and Abbeel]{ho2020denoising}
Jonathan Ho, Ajay Jain, and Pieter Abbeel.
\newblock Denoising diffusion probabilistic models.
\newblock \emph{Advances in neural information processing systems}, 33:\penalty0 6840--6851, 2020.

\bibitem[Ho et~al.(2022)Ho, Chan, Saharia, Whang, Gao, Gritsenko, Kingma, Poole, Norouzi, Fleet, et~al.]{ho2022imagen}
Jonathan Ho, William Chan, Chitwan Saharia, Jay Whang, Ruiqi Gao, Alexey Gritsenko, Diederik~P Kingma, Ben Poole, Mohammad Norouzi, David~J Fleet, et~al.
\newblock Imagen video: High definition video generation with diffusion models.
\newblock \emph{arXiv preprint arXiv:2210.02303}, 2022.

\bibitem[Islam and Bertasius(2022)]{variant1}
Md~Mohaiminul Islam and Gedas Bertasius.
\newblock Long movie clip classification with state-space video models.
\newblock In \emph{European Conference on Computer Vision}, pages 87--104. Springer, 2022.

\bibitem[Isobe et~al.(2020{\natexlab{a}})Isobe, Jia, Gu, Li, Wang, and Tian]{isobe2020video}
Takashi Isobe, Xu Jia, Shuhang Gu, Songjiang Li, Shengjin Wang, and Qi Tian.
\newblock Video super-resolution with recurrent structure-detail network.
\newblock In \emph{Computer Vision--ECCV 2020: 16th European Conference, Glasgow, UK, August 23--28, 2020, Proceedings, Part XII 16}, pages 645--660. Springer, 2020{\natexlab{a}}.

\bibitem[Isobe et~al.(2020{\natexlab{b}})Isobe, Li, Jia, Yuan, Slabaugh, Xu, Li, Wang, and Tian]{isobe2020video2}
Takashi Isobe, Songjiang Li, Xu Jia, Shanxin Yuan, Gregory Slabaugh, Chunjing Xu, Ya-Li Li, Shengjin Wang, and Qi Tian.
\newblock Video super-resolution with temporal group attention.
\newblock In \emph{Proceedings of the IEEE/CVF conference on computer vision and pattern recognition}, pages 8008--8017, 2020{\natexlab{b}}.

\bibitem[Isobe et~al.(2020{\natexlab{c}})Isobe, Zhu, Jia, and Wang]{isobe2020revisiting}
Takashi Isobe, Fang Zhu, Xu Jia, and Shengjin Wang.
\newblock Revisiting temporal modeling for video super-resolution.
\newblock \emph{arXiv preprint arXiv:2008.05765}, 2020{\natexlab{c}}.

\bibitem[Jia et~al.(2016)Jia, Brabandere, Tuytelaars, and Gool]{dynamic_filter}
Xu Jia, Bert~De Brabandere, Tinne Tuytelaars, and Luc~Van Gool.
\newblock Dynamic filter networks.
\newblock In \emph{{NIPS}}, pages 667--675, 2016.

\bibitem[Jo et~al.(2018)Jo, Oh, Kang, and Kim]{jo2018deep}
Younghyun Jo, Seoung~Wug Oh, Jaeyeon Kang, and Seon~Joo Kim.
\newblock Deep video super-resolution network using dynamic upsampling filters without explicit motion compensation.
\newblock In \emph{Proceedings of the IEEE conference on computer vision and pattern recognition}, pages 3224--3232, 2018.

\bibitem[Kalman(1960)]{kalman1960new}
Rudolph~Emil Kalman.
\newblock A new approach to linear filtering and prediction problems.
\newblock 1960.

\bibitem[Kawar et~al.(2022)Kawar, Elad, Ermon, and Song]{kawar2022denoising}
Bahjat Kawar, Michael Elad, Stefano Ermon, and Jiaming Song.
\newblock Denoising diffusion restoration models.
\newblock \emph{Advances in Neural Information Processing Systems}, 35:\penalty0 23593--23606, 2022.

\bibitem[Kawar et~al.(2023)Kawar, Zada, Lang, Tov, Chang, Dekel, Mosseri, and Irani]{kawar2023imagic}
Bahjat Kawar, Shiran Zada, Oran Lang, Omer Tov, Huiwen Chang, Tali Dekel, Inbar Mosseri, and Michal Irani.
\newblock Imagic: Text-based real image editing with diffusion models.
\newblock In \emph{Proceedings of the IEEE/CVF Conference on Computer Vision and Pattern Recognition}, pages 6007--6017, 2023.

\bibitem[Ke et~al.(2021)Ke, Wang, Wang, Milanfar, and Yang]{MUSIQ}
Junjie Ke, Qifei Wang, Yilin Wang, Peyman Milanfar, and Feng Yang.
\newblock Musiq: Multi-scale image quality transformer.
\newblock In \emph{Proceedings of the IEEE/CVF international conference on computer vision}, pages 5148--5157, 2021.

\bibitem[Kinga et~al.(2015)Kinga, Adam, et~al.]{kinga2015method}
D Kinga, Jimmy~Ba Adam, et~al.
\newblock A method for stochastic optimization.
\newblock In \emph{International conference on learning representations (ICLR)}, page~6. San Diego, California;, 2015.

\bibitem[Lab and etc.(2024)]{pku_yuan_lab_and_tuzhan_ai_etc_2024_10948109}
PKU-Yuan Lab and Tuzhan~AI etc.
\newblock Open-sora-plan, 2024.

\bibitem[Lai et~al.(2018)Lai, Huang, Wang, Shechtman, Yumer, and Yang]{lai2018learningwe}
Wei-Sheng Lai, Jia-Bin Huang, Oliver Wang, Eli Shechtman, Ersin Yumer, and Ming-Hsuan Yang.
\newblock Learning blind video temporal consistency.
\newblock In \emph{ECCV}, pages 170--185, 2018.

\bibitem[Li et~al.(2024)Li, Singh, and Grover]{domain1}
Shufan Li, Harkanwar Singh, and Aditya Grover.
\newblock Mamba-nd: Selective state space modeling for multi-dimensional data.
\newblock \emph{arXiv preprint arXiv:2402.05892}, 2024.

\bibitem[Li et~al.(2020)Li, Tao, Guo, Qi, Lu, and Jia]{li2020mucan}
Wenbo Li, Xin Tao, Taian Guo, Lu Qi, Jiangbo Lu, and Jiaya Jia.
\newblock Mucan: Multi-correspondence aggregation network for video super-resolution.
\newblock In \emph{Computer Vision--ECCV 2020: 16th European Conference, Glasgow, UK, August 23--28, 2020, Proceedings, Part X 16}, pages 335--351. Springer, 2020.

\bibitem[Liang et~al.(2024{\natexlab{a}})Liang, Zhou, Wang, Zhu, Xu, Zou, Ye, and Bai]{domain2}
Dingkang Liang, Xin Zhou, Xinyu Wang, Xingkui Zhu, Wei Xu, Zhikang Zou, Xiaoqing Ye, and Xiang Bai.
\newblock Pointmamba: A simple state space model for point cloud analysis.
\newblock \emph{arXiv preprint arXiv:2402.10739}, 2024{\natexlab{a}}.

\bibitem[Liang et~al.(2022)Liang, Fan, Xiang, Ranjan, Ilg, Green, Cao, Zhang, Timofte, and Gool]{liang2022recurrent}
Jingyun Liang, Yuchen Fan, Xiaoyu Xiang, Rakesh Ranjan, Eddy Ilg, Simon Green, Jiezhang Cao, Kai Zhang, Radu Timofte, and Luc~V Gool.
\newblock Recurrent video restoration transformer with guided deformable attention.
\newblock \emph{Advances in Neural Information Processing Systems}, 35:\penalty0 378--393, 2022.

\bibitem[Liang et~al.(2024{\natexlab{b}})Liang, Cao, Fan, Zhang, Ranjan, Li, Timofte, and Van~Gool]{liang2024vrt}
Jingyun Liang, Jiezhang Cao, Yuchen Fan, Kai Zhang, Rakesh Ranjan, Yawei Li, Radu Timofte, and Luc Van~Gool.
\newblock Vrt: A video restoration transformer.
\newblock \emph{IEEE Transactions on Image Processing}, 2024{\natexlab{b}}.

\bibitem[Liu and Sun(2013)]{liu2013bayesian}
Ce Liu and Deqing Sun.
\newblock On bayesian adaptive video super resolution.
\newblock \emph{IEEE transactions on pattern analysis and machine intelligence}, 36\penalty0 (2):\penalty0 346--360, 2013.

\bibitem[Lugmayr et~al.(2022)Lugmayr, Danelljan, Romero, Yu, Timofte, and Van~Gool]{lugmayr2022repaint}
Andreas Lugmayr, Martin Danelljan, Andres Romero, Fisher Yu, Radu Timofte, and Luc Van~Gool.
\newblock Repaint: Inpainting using denoising diffusion probabilistic models.
\newblock In \emph{Proceedings of the IEEE/CVF conference on computer vision and pattern recognition}, pages 11461--11471, 2022.

\bibitem[Mittal et~al.(2012)Mittal, Soundararajan, and Bovik]{NIQE}
Anish Mittal, Rajiv Soundararajan, and Alan~C Bovik.
\newblock Making a “completely blind” image quality analyzer.
\newblock \emph{IEEE Signal processing letters}, 20\penalty0 (3):\penalty0 209--212, 2012.

\bibitem[Nah et~al.(2019)Nah, Baik, Hong, Moon, Son, Timofte, and Mu~Lee]{nah2019ntire}
Seungjun Nah, Sungyong Baik, Seokil Hong, Gyeongsik Moon, Sanghyun Son, Radu Timofte, and Kyoung Mu~Lee.
\newblock Ntire 2019 challenge on video deblurring and super-resolution: Dataset and study.
\newblock In \emph{Proceedings of the IEEE/CVF conference on computer vision and pattern recognition workshops}, pages 0--0, 2019.

\bibitem[Nguyen et~al.(2022)Nguyen, Goel, Gu, Downs, Shah, Dao, Baccus, and R{\'e}]{variant2}
Eric Nguyen, Karan Goel, Albert Gu, Gordon Downs, Preey Shah, Tri Dao, Stephen Baccus, and Christopher R{\'e}.
\newblock S4nd: Modeling images and videos as multidimensional signals with state spaces.
\newblock \emph{Advances in neural information processing systems}, 35:\penalty0 2846--2861, 2022.

\bibitem[Pan et~al.(2021)Pan, Bai, Dong, Zhang, and Tang]{DBVSR}
Jinshan Pan, Haoran Bai, Jiangxin Dong, Jiawei Zhang, and Jinhui Tang.
\newblock Deep blind video super-resolution.
\newblock In \emph{Proceedings of the IEEE/CVF International Conference on Computer Vision}, pages 4811--4820, 2021.

\bibitem[Polyak et~al.(2024)Polyak, Zohar, Brown, Tjandra, Sinha, Lee, Vyas, Shi, Ma, Chuang, et~al.]{polyak2024movie}
Adam Polyak, Amit Zohar, Andrew Brown, Andros Tjandra, Animesh Sinha, Ann Lee, Apoorv Vyas, Bowen Shi, Chih-Yao Ma, Ching-Yao Chuang, et~al.
\newblock Movie gen: A cast of media foundation models.
\newblock \emph{arXiv preprint arXiv:2410.13720}, 2024.

\bibitem[Ramesh et~al.(2022)Ramesh, Dhariwal, Nichol, Chu, and Chen]{ramesh2022hierarchical}
Aditya Ramesh, Prafulla Dhariwal, Alex Nichol, Casey Chu, and Mark Chen.
\newblock Hierarchical text-conditional image generation with clip latents.
\newblock \emph{arXiv preprint arXiv:2204.06125}, 1\penalty0 (2):\penalty0 3, 2022.

\bibitem[Rombach et~al.(2022)Rombach, Blattmann, Lorenz, Esser, and Ommer]{rombach2022high}
Robin Rombach, Andreas Blattmann, Dominik Lorenz, Patrick Esser, and Bj{\"o}rn Ommer.
\newblock High-resolution image synthesis with latent diffusion models.
\newblock In \emph{Proceedings of the IEEE/CVF conference on computer vision and pattern recognition}, pages 10684--10695, 2022.

\bibitem[Saharia et~al.(2022)Saharia, Chan, Saxena, Li, Whang, Denton, Ghasemipour, Gontijo~Lopes, Karagol~Ayan, Salimans, et~al.]{saharia2022photorealistic}
Chitwan Saharia, William Chan, Saurabh Saxena, Lala Li, Jay Whang, Emily~L Denton, Kamyar Ghasemipour, Raphael Gontijo~Lopes, Burcu Karagol~Ayan, Tim Salimans, et~al.
\newblock Photorealistic text-to-image diffusion models with deep language understanding.
\newblock \emph{Advances in neural information processing systems}, 35:\penalty0 36479--36494, 2022.

\bibitem[Smith et~al.(2022)Smith, Warrington, and Linderman]{variant5}
Jimmy~TH Smith, Andrew Warrington, and Scott~W Linderman.
\newblock Simplified state space layers for sequence modeling.
\newblock \emph{arXiv preprint arXiv:2208.04933}, 2022.

\bibitem[Tao et~al.(2017)Tao, Gao, Liao, Wang, and Jia]{tao2017detail}
Xin Tao, Hongyun Gao, Renjie Liao, Jue Wang, and Jiaya Jia.
\newblock Detail-revealing deep video super-resolution.
\newblock In \emph{Proceedings of the IEEE international conference on computer vision}, pages 4472--4480, 2017.

\bibitem[Wang et~al.(2023{\natexlab{a}})Wang, Chan, and Loy]{wang2023exploring}
Jianyi Wang, Kelvin~CK Chan, and Chen~Change Loy.
\newblock Exploring clip for assessing the look and feel of images.
\newblock In \emph{Proceedings of the AAAI Conference on Artificial Intelligence}, pages 2555--2563, 2023{\natexlab{a}}.

\bibitem[Wang et~al.(2023{\natexlab{b}})Wang, Zhu, Wang, Yu, Liu, Omar, and Hamid]{variant3}
Jue Wang, Wentao Zhu, Pichao Wang, Xiang Yu, Linda Liu, Mohamed Omar, and Raffay Hamid.
\newblock Selective structured state-spaces for long-form video understanding.
\newblock In \emph{Proceedings of the IEEE/CVF Conference on Computer Vision and Pattern Recognition}, pages 6387--6397, 2023{\natexlab{b}}.

\bibitem[Wang et~al.(2024{\natexlab{a}})Wang, Yue, Zhou, Chan, and Loy]{stablesr}
Jianyi Wang, Zongsheng Yue, Shangchen Zhou, Kelvin~CK Chan, and Chen~Change Loy.
\newblock Exploiting diffusion prior for real-world image super-resolution.
\newblock \emph{International Journal of Computer Vision}, pages 1--21, 2024{\natexlab{a}}.

\bibitem[Wang et~al.(2024{\natexlab{b}})Wang, Yue, Zhou, Chan, and Loy]{wang2024exploiting}
Jianyi Wang, Zongsheng Yue, Shangchen Zhou, Kelvin~CK Chan, and Chen~Change Loy.
\newblock Exploiting diffusion prior for real-world image super-resolution.
\newblock \emph{International Journal of Computer Vision}, pages 1--21, 2024{\natexlab{b}}.

\bibitem[Wang et~al.(2019)Wang, Chan, Yu, Dong, and Change~Loy]{wang2019edvr}
Xintao Wang, Kelvin~CK Chan, Ke Yu, Chao Dong, and Chen Change~Loy.
\newblock Edvr: Video restoration with enhanced deformable convolutional networks.
\newblock In \emph{Proceedings of the IEEE/CVF conference on computer vision and pattern recognition workshops}, pages 0--0, 2019.

\bibitem[Wang et~al.(2021)Wang, Xie, Dong, and Shan]{RealESRGAN}
Xintao Wang, Liangbin Xie, Chao Dong, and Ying Shan.
\newblock Real-esrgan: Training real-world blind super-resolution with pure synthetic data.
\newblock In \emph{Proceedings of the IEEE/CVF international conference on computer vision}, pages 1905--1914, 2021.

\bibitem[Wu et~al.(2023)Wu, Zhang, Liao, Chen, Hou, Wang, Sun, Yan, and Lin]{DOVER}
Haoning Wu, Erli Zhang, Liang Liao, Chaofeng Chen, Jingwen Hou, Annan Wang, Wenxiu Sun, Qiong Yan, and Weisi Lin.
\newblock Exploring video quality assessment on user generated contents from aesthetic and technical perspectives.
\newblock In \emph{Proceedings of the IEEE/CVF International Conference on Computer Vision}, pages 20144--20154, 2023.

\bibitem[Xie et~al.(2023)Xie, Wang, Shi, Gu, Dong, and Shan]{xie2023mitigating}
Liangbin Xie, Xintao Wang, Shuwei Shi, Jinjin Gu, Chao Dong, and Ying Shan.
\newblock Mitigating artifacts in real-world video super-resolution models.
\newblock In \emph{Proceedings of the AAAI Conference on Artificial Intelligence}, pages 2956--2964, 2023.

\bibitem[Xie et~al.(2022)Xie, Zhang, Cao, Lin, Bao, Yao, Dai, and Hu]{self-supervised_task8}
Zhenda Xie, Zheng Zhang, Yue Cao, Yutong Lin, Jianmin Bao, Zhuliang Yao, Qi Dai, and Han Hu.
\newblock Simmim: A simple framework for masked image modeling.
\newblock In \emph{Proceedings of the IEEE/CVF conference on computer vision and pattern recognition}, pages 9653--9663, 2022.

\bibitem[Xing et~al.(2024)Xing, Ye, Yang, Liu, and Zhu]{xing2024segmamba}
Zhaohu Xing, Tian Ye, Yijun Yang, Guang Liu, and Lei Zhu.
\newblock Segmamba: Long-range sequential modeling mamba for 3d medical image segmentation.
\newblock In \emph{International Conference on Medical Image Computing and Computer-Assisted Intervention}, pages 578--588. Springer, 2024.

\bibitem[Xue et~al.(2019)Xue, Chen, Wu, Wei, and Freeman]{xue2019video}
Tianfan Xue, Baian Chen, Jiajun Wu, Donglai Wei, and William~T Freeman.
\newblock Video enhancement with task-oriented flow.
\newblock \emph{International Journal of Computer Vision}, 127:\penalty0 1106--1125, 2019.

\bibitem[Yang et~al.(2023{\natexlab{a}})Yang, Wu, Ren, Xie, and Zhang]{yang2023pixel}
Tao Yang, Rongyuan Wu, Peiran Ren, Xuansong Xie, and Lei Zhang.
\newblock Pixel-aware stable diffusion for realistic image super-resolution and personalized stylization.
\newblock \emph{arXiv preprint arXiv:2308.14469}, 2023{\natexlab{a}}.

\bibitem[Yang et~al.(2021)Yang, Xiang, Zeng, and Zhang]{realvsr}
Xi Yang, Wangmeng Xiang, Hui Zeng, and Lei Zhang.
\newblock Real-world video super-resolution: A benchmark dataset and a decomposition based learning scheme.
\newblock In \emph{Proceedings of the IEEE/CVF International Conference on Computer Vision}, pages 4781--4790, 2021.

\bibitem[Yang et~al.(2023{\natexlab{b}})Yang, He, Ma, and Zhang]{mgld}
Xi Yang, Chenhang He, Jianqi Ma, and Lei Zhang.
\newblock Motion-guided latent diffusion for temporally consistent real-world video super-resolution.
\newblock \emph{arXiv preprint arXiv:2312.00853}, 2023{\natexlab{b}}.

\bibitem[Yi et~al.(2019)Yi, Wang, Jiang, Jiang, and Ma]{yi2019progressive}
Peng Yi, Zhongyuan Wang, Kui Jiang, Junjun Jiang, and Jiayi Ma.
\newblock Progressive fusion video super-resolution network via exploiting non-local spatio-temporal correlations.
\newblock In \emph{Proceedings of the IEEE/CVF international conference on computer vision}, pages 3106--3115, 2019.

\bibitem[Zhang et~al.(2023)Zhang, Rao, and Agrawala]{controlnet}
Lvmin Zhang, Anyi Rao, and Maneesh Agrawala.
\newblock Adding conditional control to text-to-image diffusion models, 2023.

\bibitem[Zhang et~al.(2018)Zhang, Isola, Efros, Shechtman, and Wang]{LPIPS}
Richard Zhang, Phillip Isola, Alexei~A Efros, Eli Shechtman, and Oliver Wang.
\newblock The unreasonable effectiveness of deep features as a perceptual metric.
\newblock In \emph{Proceedings of the IEEE conference on computer vision and pattern recognition}, pages 586--595, 2018.

\bibitem[Zhang and Yao(2024)]{zhang2024realviformer}
Yuehan Zhang and Angela Yao.
\newblock Realviformer: Investigating attention for real-world video super-resolution.
\newblock \emph{arXiv preprint arXiv:2407.13987}, 2024.

\bibitem[Zhou et~al.(2024)Zhou, Yang, Wang, Luo, and Loy]{upscale}
Shangchen Zhou, Peiqing Yang, Jianyi Wang, Yihang Luo, and Chen~Change Loy.
\newblock Upscale-a-video: Temporal-consistent diffusion model for real-world video super-resolution.
\newblock In \emph{Proceedings of the IEEE/CVF Conference on Computer Vision and Pattern Recognition}, pages 2535--2545, 2024.

\bibitem[Zhu et~al.(2024)Zhu, Liao, Zhang, Wang, Liu, and Wang]{domain3}
Lianghui Zhu, Bencheng Liao, Qian Zhang, Xinlong Wang, Wenyu Liu, and Xinggang Wang.
\newblock Vision mamba: Efficient visual representation learning with bidirectional state space model.
\newblock \emph{arXiv preprint arXiv:2401.09417}, 2024.

\end{thebibliography}
